\documentclass[letterpaper]{article} 
\usepackage{aaai2026}  
\usepackage{times}  
\usepackage{helvet}  
\usepackage{courier}  
\usepackage[hyphens]{url}  
\usepackage{graphicx} 
\usepackage{natbib}  
\usepackage{caption} 
\usepackage{mwe}
\usepackage{graphicx}
\usepackage{amsmath}
\usepackage{algorithm}
\usepackage{algpseudocode}
\usepackage{subcaption} 
\usepackage{tabularray}
\usepackage{booktabs}

\usepackage{newfloat}
\usepackage{listings}
\DeclareCaptionStyle{ruled}{labelfont=normalfont,labelsep=colon,strut=off} 
\floatstyle{ruled}
\newfloat{listing}{tb}{lst}{}
\floatname{listing}{Listing}
\title{Bridging Streaming Continual Learning via In-Context Large Tabular Models}
\author{
    Afonso Lourenço\textsuperscript{\rm 1}, João Gama\textsuperscript{\rm 2}, Eric P. Xing\textsuperscript{\rm 3,4}, Goreti Marreiros\textsuperscript{\rm 1}}
\affiliations{
    \textsuperscript{\rm 1}GECAD, ISEP, Polytechnic of Porto, Rua Dr. António Bernardino de Almeida, Porto, 4249-015, Portugal\\
    \textsuperscript{\rm 2}INESC-TEC, FEP, University of Porto, Rua Dr. Roberto Frias, Porto, 4200-465, Portugal \\
    \textsuperscript{\rm 3}Mohamed bin Zayed University of Artificial Intelligence, Abu Dhabi, UAE \\
    \textsuperscript{\rm 4}Carnegie Mellon University, Pittsburgh, PA, USA
    
}

\usepackage{bibentry}

\begin{document}

\nocopyright 

\maketitle

\begin{abstract}
In streaming scenarios, models must learn continuously, adapting to concept drifts without erasing previously acquired knowledge. However, existing research communities address these challenges in isolation. Continual Learning (CL) focuses on long-term retention and mitigating catastrophic forgetting, often without strict real-time constraints. Stream Learning (SL) emphasizes rapid, efficient adaptation to high-frequency data streams, but typically neglects forgetting. Recent efforts have tried to combine these paradigms, yet no clear algorithmic overlap exists. We argue that large in-context tabular models (LTMs) provide a natural bridge for Streaming Continual Learning (SCL). In our view, unbounded streams should be summarized on-the-fly into compact sketches that can be consumed by LTMs. This recovers the classical SL motivation of compressing massive streams with fixed-size guarantees, while simultaneously aligning with the experience-replay desiderata of CL. To clarify this bridge, we show how the SL and CL communities implicitly adopt a divide-to-conquer strategy to manage the tension between plasticity (performing well on the current distribution) and stability (retaining past knowledge), while also imposing a minimal complexity constraint that motivates diversification (avoiding redundancy in what is stored) and retrieval (re-prioritizing past information when needed). Within this perspective, we propose structuring SCL with LTMs around two core principles of data selection for in-context learning: (1) distribution matching, which balances plasticity and stability, and (2) distribution compression, which controls memory size through diversification and retrieval mechanisms.
\end{abstract}

\section{Introduction}

\begin{figure}[ht]
    \centering
    \begin{subfigure}[t]{0.17\textwidth}
        \centering
        \includegraphics[width=\textwidth]{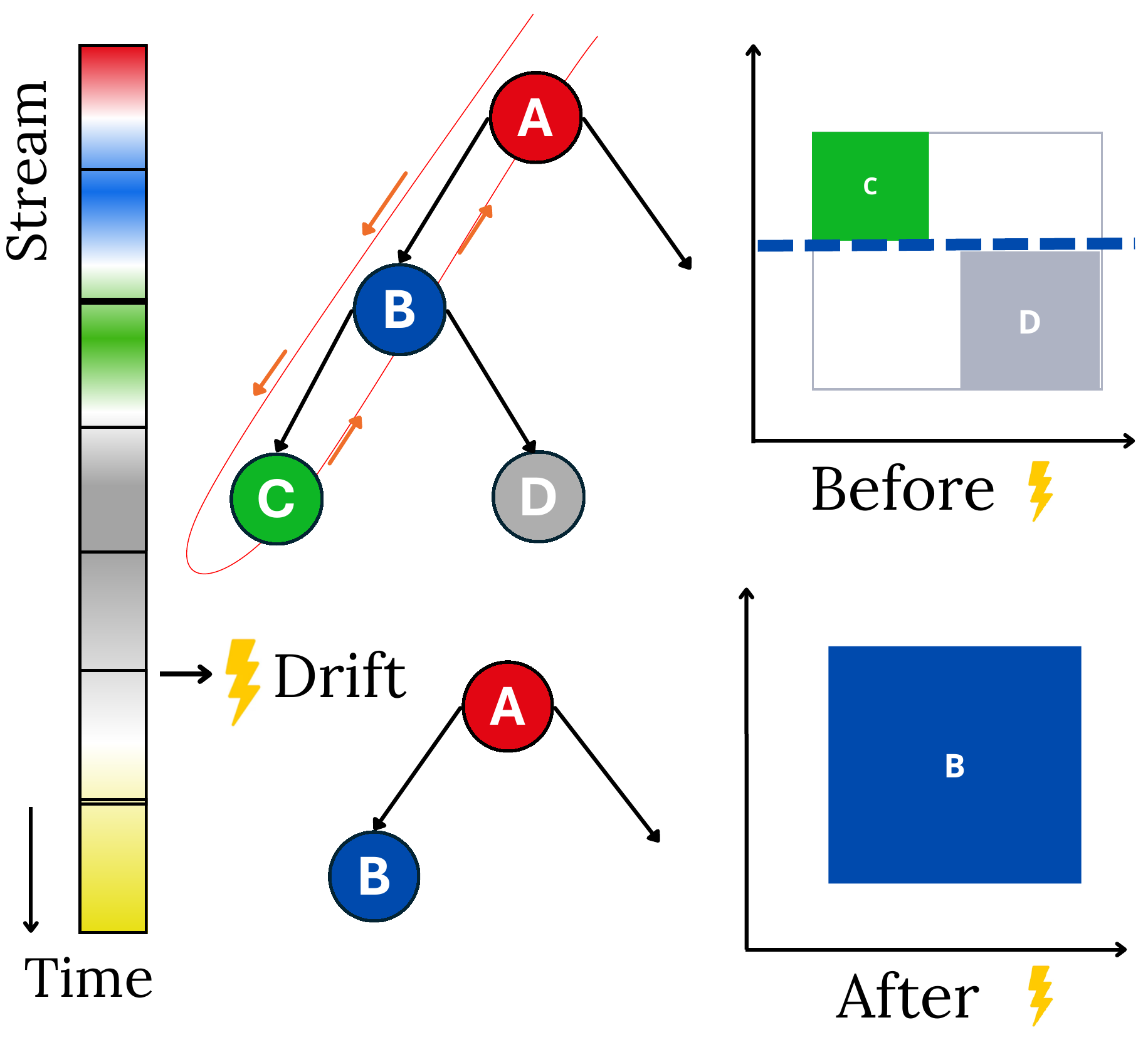}
        \caption{IDTs.}
        \label{fig:idts1}
    \end{subfigure}
    \hfill
    \begin{subfigure}[t]{0.29\textwidth}
        \centering
        \includegraphics[width=\textwidth]{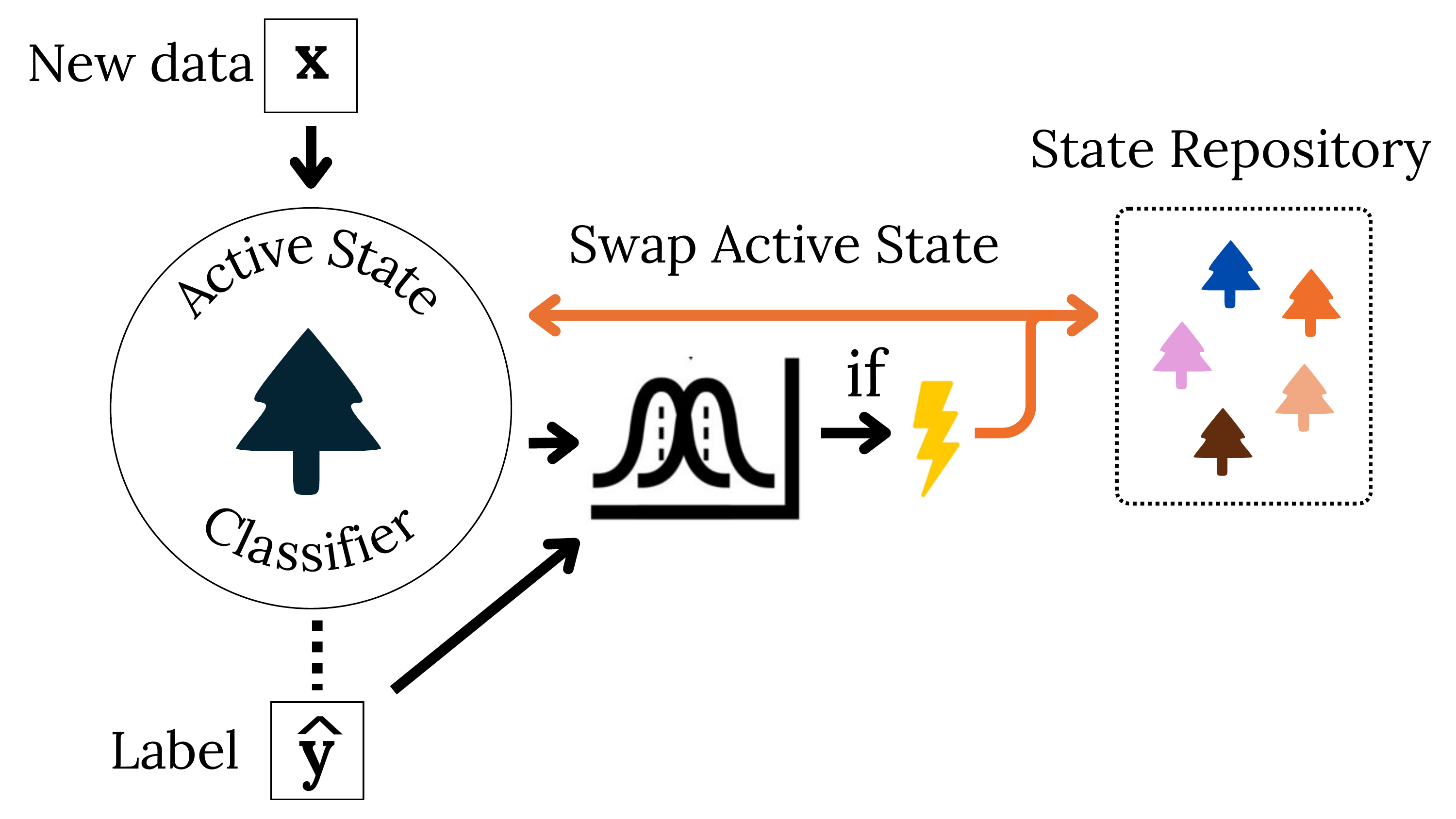}
        \caption{IDTs swapped upon drift alarm.}
        \label{fig:idts2}
    \end{subfigure}
    \caption{SL relies on structural expansion, mostly adding new intelligence components for ensembling of IDTs.}
\end{figure}

For tabular stream learning (SL), ensembles of incremental decision trees (IDTs) have long been state-of-the-art \cite{krawczyk2017ensemble}. They use statistical bounds to decide node splits and handle concept drift via subtree replacement (Fig. \ref{fig:idts1}). As shallow learners, IDTs converge quickly online due to their few trainable parameters. Yet, their learning capacity is limited by single-view features, plasticity loss from making locally optimal splits, catastrophic forgetting of class-conditional estimators, and the inability to model dependencies. Although various ad-hoc solutions have been proposed, mostly adding new candidate components for ensembling, they tend to be narrow, addressing one problem while assuming others are controlled. For example, ensembles may use drift detectors to swap to a more suitable model (Fig. \ref{fig:idts2}), but fail to evaluate stored models for relevance if the drift does not trigger an alarm \cite{halstead2022probabilistic}.

\begin{figure}[ht]
    \centering
    \includegraphics[width=0.39\textwidth]{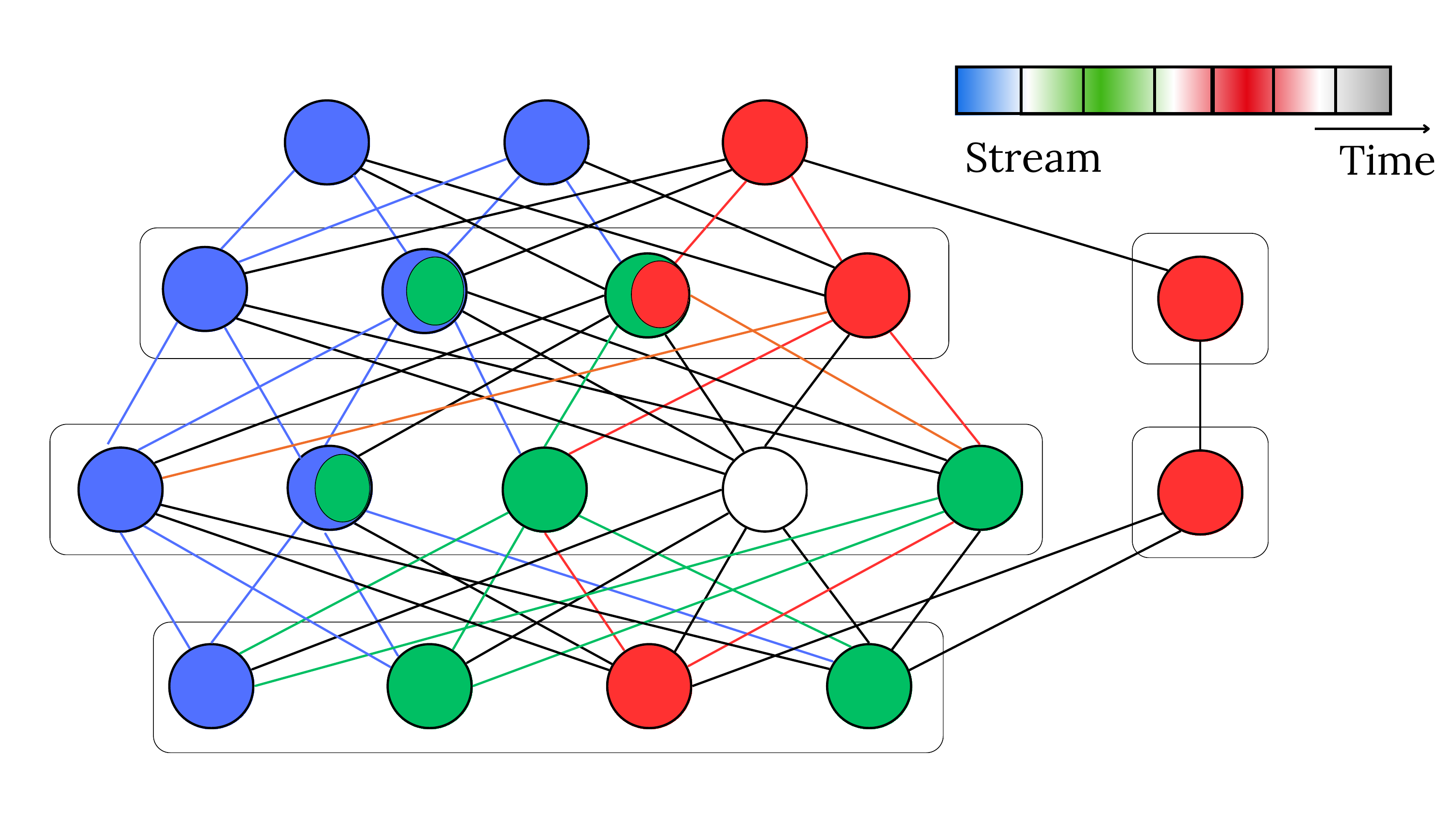}
    \caption{CL not only relies on structural expansion, but also sophisticated parameter adaptation/activation DL schemas.}
    \label{fig:dl}
\end{figure}

Indeed, these ad-hoc intelligence components are only useful if we integrate them into a truly autonomous system. In this regard, continual learning (CL) allows more powerful deep learning (DL) schemas, adapting both through parameter addition and activation (Fig. \ref{fig:dl}). However, these struggle with tabular streams \cite{sahoo2017online}. On one hand, the inductive biases of DL architectures assume structures which offer little advantage for irregular patterns typical of tabular data \cite{mcelfresh2023neural}. On the other hand, entangled architectures converge slowly due to stochastic updates, and unfixed weights make prior knowledge prone to being overwritten. Moreover, this plasticity does not guarantee learning efficiency, often requiring multiple data passes to reduce interference \cite{lyle2023understanding}.

\begin{figure}[htbp]  
    \centering
    \begin{subfigure}[t]{0.23\textwidth}
        \centering
        \includegraphics[width=\textwidth]{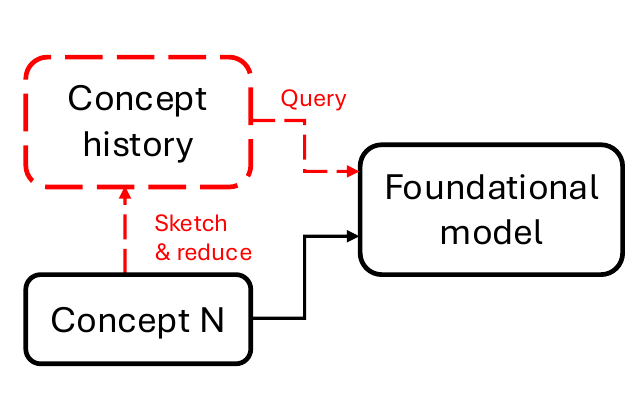}
        \caption{In-context stream mining.}
    \end{subfigure}
    \begin{subfigure}[t]{0.23\textwidth}
        \centering
        \includegraphics[width=\textwidth]{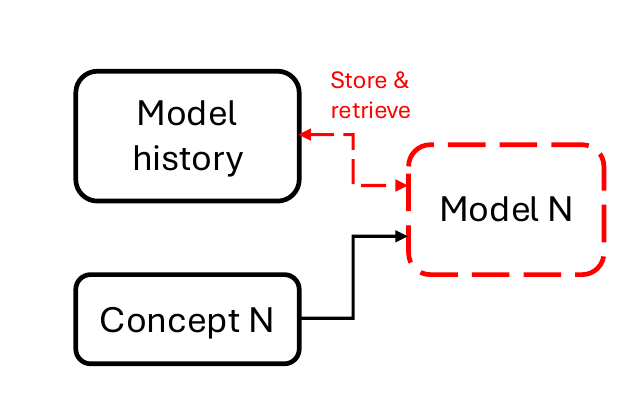}
        \caption{Conventional SL and CL.}
    \end{subfigure}
    \caption{While SL and CL methods adapt model parameters, SCL via LTMs instead treats input data as the learnable object and keeps model parameters fixed. Red = Trainable.}
    \label{fig:newparadigm}
\end{figure}

Due to these algorithmic differences, SL and CL prioritize different aspects of stateful learning \cite{lourencco2025device}. CL focuses on long-term retention and mitigating forgetting, often without strict real-time constraints \cite{lin2023theory}, whereas SL emphasizes rapid adaptation to high-frequency streams but typically ignores high-order dependencies and forgetting \cite{gama2010knowledge}. Recent efforts have tried to combine these paradigms \cite{gunasekara2023survey,giannini2024streaming}, yet no clear overlap exists. To address this, we propose to leverage the disruptive success of in-context large tabular models (LTMs) \cite{hollmann2025accurate} as the unifying bridge for streaming continual learning (SCL): using on-the-fly techniques to summarize unbounded data streams before feeding them to LTMs (Fig. \ref{fig:newparadigm}). 

In prior work \cite{lourencco2025context}, we showed that augmenting the TabPFN transformer \cite{hollmann2025accurate} with a simple inference-time sketching mechanism consistently outperforms state-of-the-art methods such as Adaptive Random Forest \cite{gomes2017adaptive}, and Streaming Random Patches \cite{gomes2019streaming}, on standard streaming benchmarks: NOAA, SmartMeter, Electricity, Rialto, Posture, CoverType, and PokerHand. Here, we explain how this data-centric view unifies the strengths of SL and CL: recovering the classical SL goal of compressing massive streams into compact sketches whose size and computational cost remain bounded \cite{cormode2005improved}, while simultaneously aligning with CL’s experience-replay desiderata for retaining past concepts \cite{riemer2019learning}. Our argument unfolds in two steps:

\begin{itemize}
\item \textbf{Extracting the shared desiderata of SL and CL.} Both communities follow a divide-to-conquer strategy driven by the underlying tension between \textbf{plasticity} (adapting to the current distribution) and \textbf{stability} (retaining past knowledge). Under a minimal memory constraint, this tension gives rise to two operational requirements: \textbf{diversification}, to avoid redundant or overlapping stored information, and \textbf{retrieval}, to re-activate relevant past experience when needed.

\item \textbf{Mapping these desiderata to in-context stream mining with LTMs.} We show that SCL can be framed as selecting and organizing data for in-context learning, where \textbf{distribution matching} governs the plasticity–stability balance, and \textbf{distribution compression} governs memory efficiency through diversification and retrieval.
\end{itemize}

\section{Current SL and CL state-of-the-art}

\begin{figure}[ht]
    \centering
    \begin{subfigure}[t]{0.33\textwidth}
        \centering
        \includegraphics[width=\textwidth]{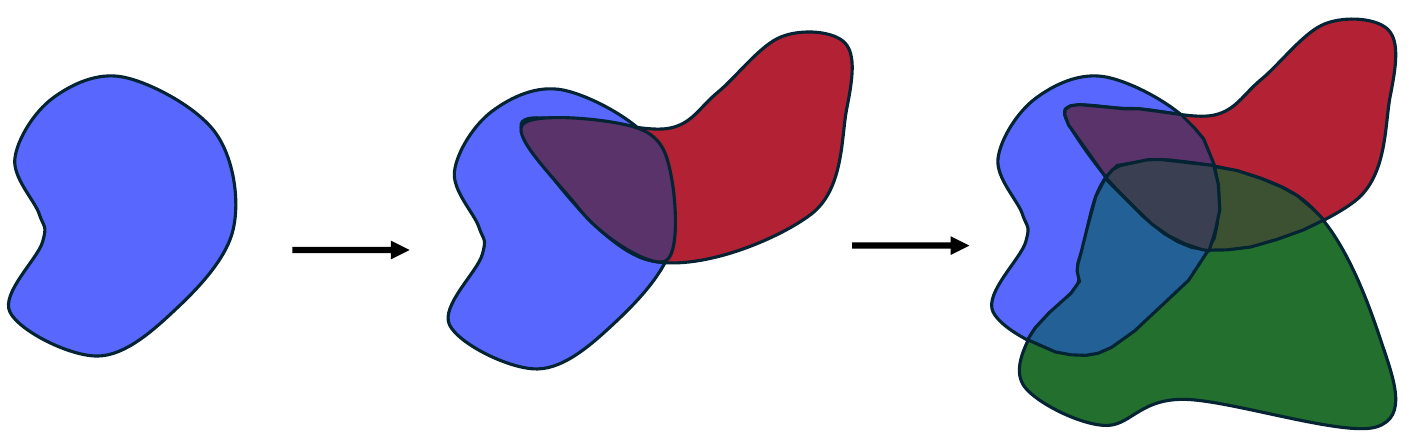}
        \caption{Incrementally constrained solution space.}
        \label{fig:NPhard}
    \end{subfigure}
    \hfill
    \begin{subfigure}[t]{0.11\textwidth}
        \centering
        \includegraphics[width=\textwidth]{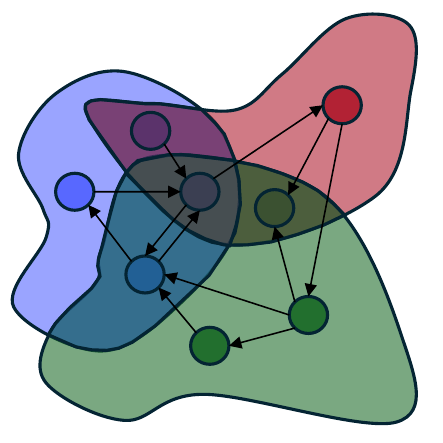}
        \caption{Modularity.}
        \label{fig:modularity}
    \end{subfigure}
    \caption{On the need for divide-to-conquer in SL and CL.}
\end{figure}

Both SL and CL ultimately seek to maintain useful knowledge over time while adapting to new data. However, when learning occurs under streaming, non-stationary conditions, this requires a divide-to-conquer strategy. As new concepts accumulate, the feasible parameter space becomes progressively restricted, making it increasingly difficult to adjust the model without interfering with previously acquired knowledge (Fig.~\ref{fig:NPhard}). In the limit, finding parameter configurations that jointly satisfy all concepts is NP-hard \cite{knoblauch2020optimal}. Thus, the challenge is not merely retaining information, but doing so while preserving efficient access and reuse. A central lever in controlling interference is the degree of parameter sharing. Full sharing maximizes generalization but risks interference; no sharing avoids interference but scales poorly with the number of concepts \cite{risca2025continual}. A middle ground is modular compositionality, where knowledge is distributed across components that can be selectively re-used, enabling forward and backward transfer (Fig.~\ref{fig:modularity}).

For tabular data streams, ensembles of IDTs are widely used and highly effective \cite{krawczyk2017ensemble}. A single IDT can adapt online and converge to near-optimal splits using one-pass summary statistics and statistically grounded decision criteria, such as information gain tested with Hoeffding-style bounds \cite{lourencco2025dfdt}. However, predictive performance emerges primarily at the ensemble level. When concept drift occurs, trees respond differently depending on their local minima and data histories. This natural diversity can be leveraged: outdated or underperforming trees can be replaced, new trees introduced, and ensemble votes re-weighted based on recent performance. Research in this area has therefore progressed along two complementary axes. The first concerns the base learners themselves, balancing their ability to adapt to new data (\textbf{plasticity}) while retaining useful structure from the past (\textbf{stability}). The second focuses on ensemble management, ensuring that the collection of learners remains diverse enough to cover different regions of the input space (\textbf{diversification}) and that previously useful models can be reactivated when similar conditions reappear (\textbf{retrieval}).

\subsection{Stability}

\begin{figure}[ht]
    \centering
    \includegraphics[width=0.48\textwidth]{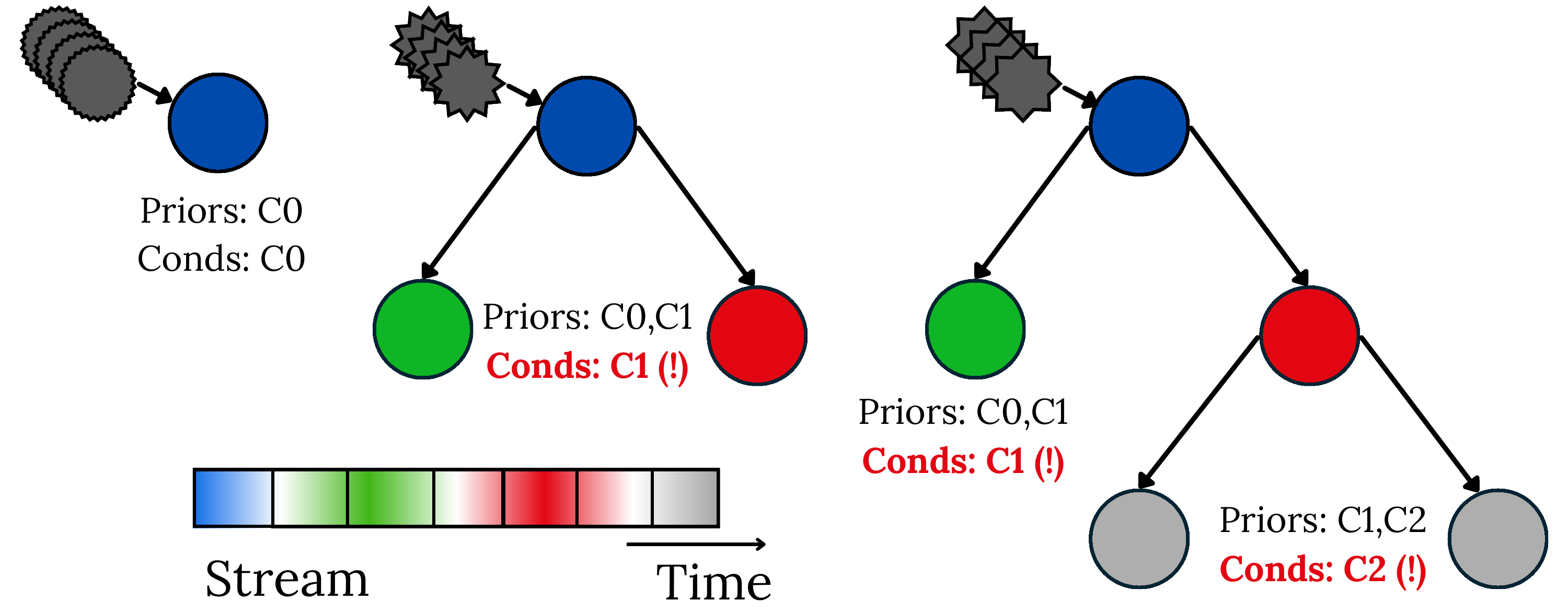}
    \caption{IDTs' conditional classification fails in class-incremental sequences, due to missing estimates.}
    \label{fig:cf}
\end{figure}

Despite their ability to continually store incrementally arriving data, IDTs are often biased toward recently observed distributions when new classes appear. Under strong temporal imbalance, where older classes do not reappear, performance on previously learned concepts deteriorates, resulting in catastrophic forgetting. This problem is amplified after splits, where conditional estimators are reset: classes absent from the current stream lose representation, are excluded from entropy-based split decisions, cannot be used in Naive Bayes classification, and have their priors removed at deeper nodes. Forgetting in IDTs occurs through three mechanisms: (1) exclusion of older classes from split evaluation, (2) failure of conditional classification due to missing estimates, and (3) disappearance of class priors in new branches (Fig. \ref{fig:cf}) \cite{korycki2021streaming}. Stability-oriented approaches mitigate this by preserving class information during updates. One strategy propagates class-conditional attribute estimators and maintains class priors in entropy and Bayesian computations \cite{korycki2021streaming}. Another relies on short-term memory replay to preserve representation continuity, e.g., through oversampling \cite{bobowska2019online}, or per-class balanced queues \cite{malialis2020online}.

\begin{figure}[ht]
    \centering
    \includegraphics[width=0.4\textwidth]{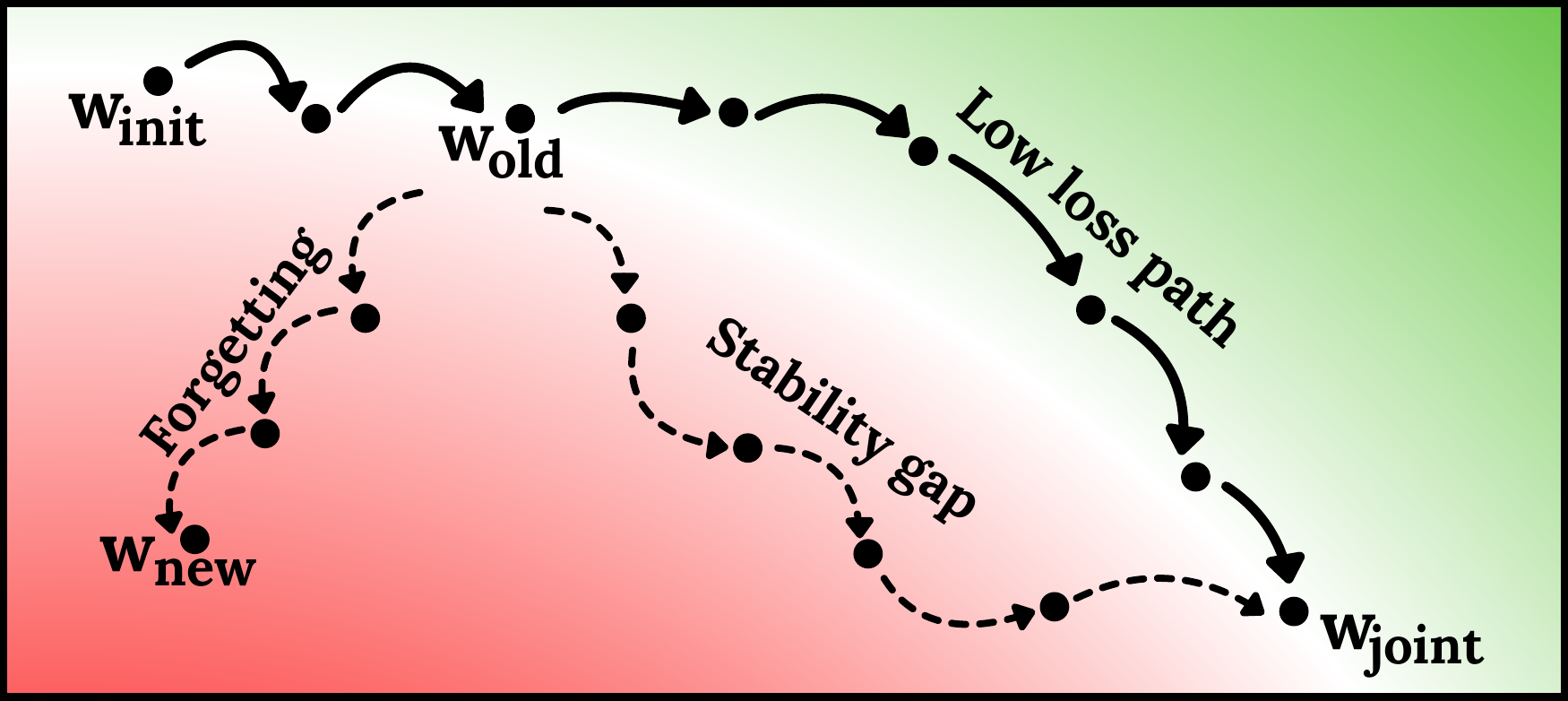}
    \caption{When optimizing the joint loss, SGD typically passes through a region of high loss on old tasks.}
    \label{fig:optimize}
\end{figure}

Neural networks allow more expressive stability mechanisms. A common strategy is to approximate joint optimization over past tasks by penalizing updates to parameters deemed important for previous ones (Fig. \ref{fig:optimize}). EWC \cite{kirkpatrick2017overcoming} measures importance via the Fisher Information Matrix, MAS \cite{aljundi2018memory} via gradient or Hebbian activity with constant memory, and SI \cite{zenke2017synaptic} accumulates importance online from loss reductions. RWalk \cite{chaudhry2018riemannian} unifies these by computing importance in the Fisher-induced Riemannian parameter space. Since all of these operate incrementally in weight space, they extend naturally to streaming settings \cite{elsayed2024addressing}. However, such methods typically exhibit temporary forgetting because SGD must pass through regions of high loss on old tasks to reach the regularized joint optimum \cite{de2022continual}. This stability gap motivates modifying not only the objective, but also the optimization path \cite{hess2023two}. To address this, gradient projection methods enforce updates orthogonal to gradient subspaces of previous tasks \cite{farajtabar2020orthogonal,guo2022adaptive}. However, while this reduces forgetting, strict orthogonality can overly limit knowledge transfer. Recent work relaxes these constraints, enabling controlled sharing, e.g., NCL \cite{kao2021natural} re-scales gradients using a Kronecker-factored Fisher approximation and combines projection with parameter regularization.

\subsection{Plasticity}

\begin{figure}[ht]
    \centering
    \begin{subfigure}[t]{0.21\textwidth}
        \centering
        \includegraphics[width=\textwidth]{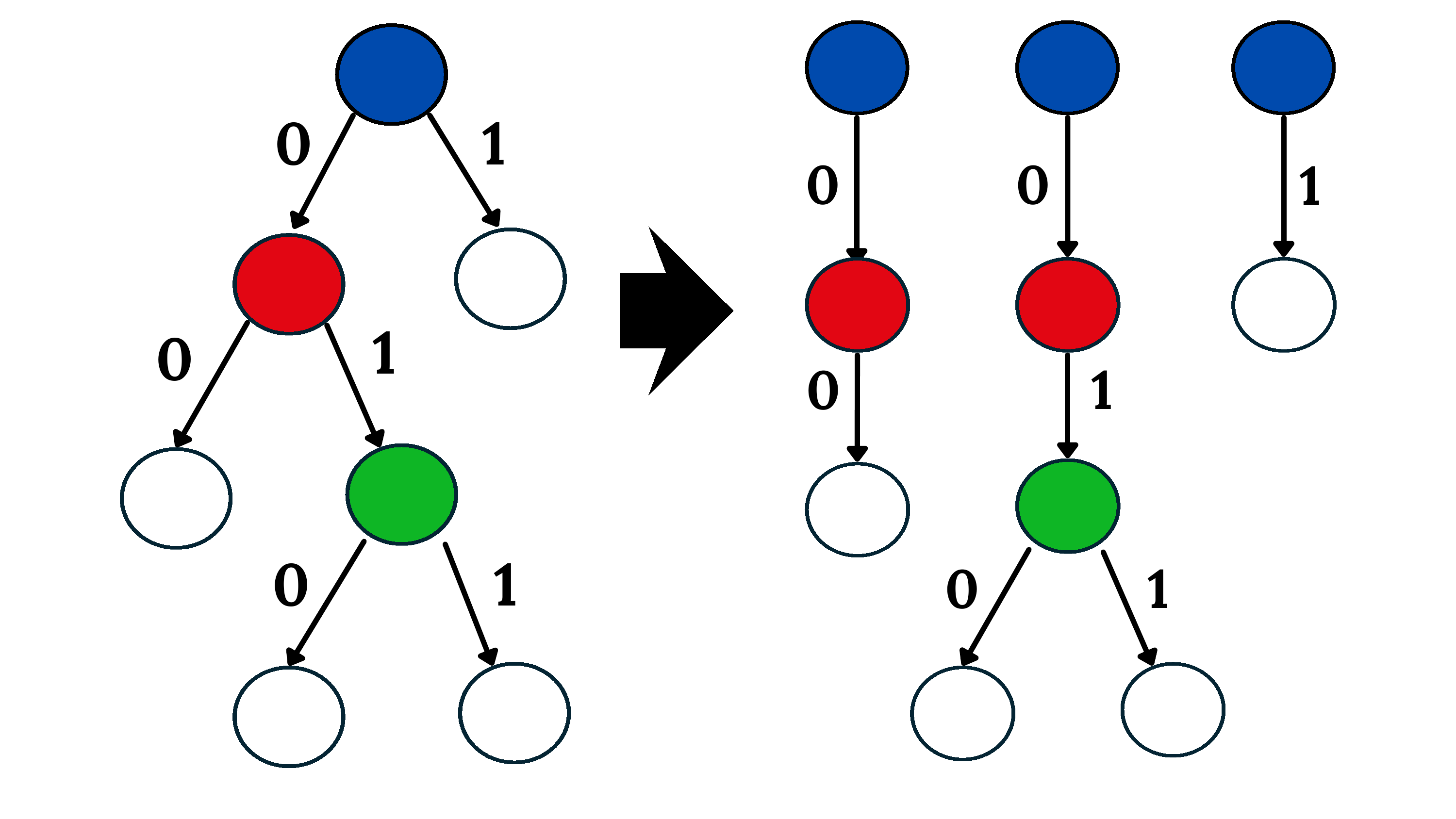}
        \caption{Disconnect subtree.}
        \label{fig:step1}
    \end{subfigure}
    \hfill
    \begin{subfigure}[t]{0.24\textwidth}
        \centering
        \includegraphics[width=\textwidth]{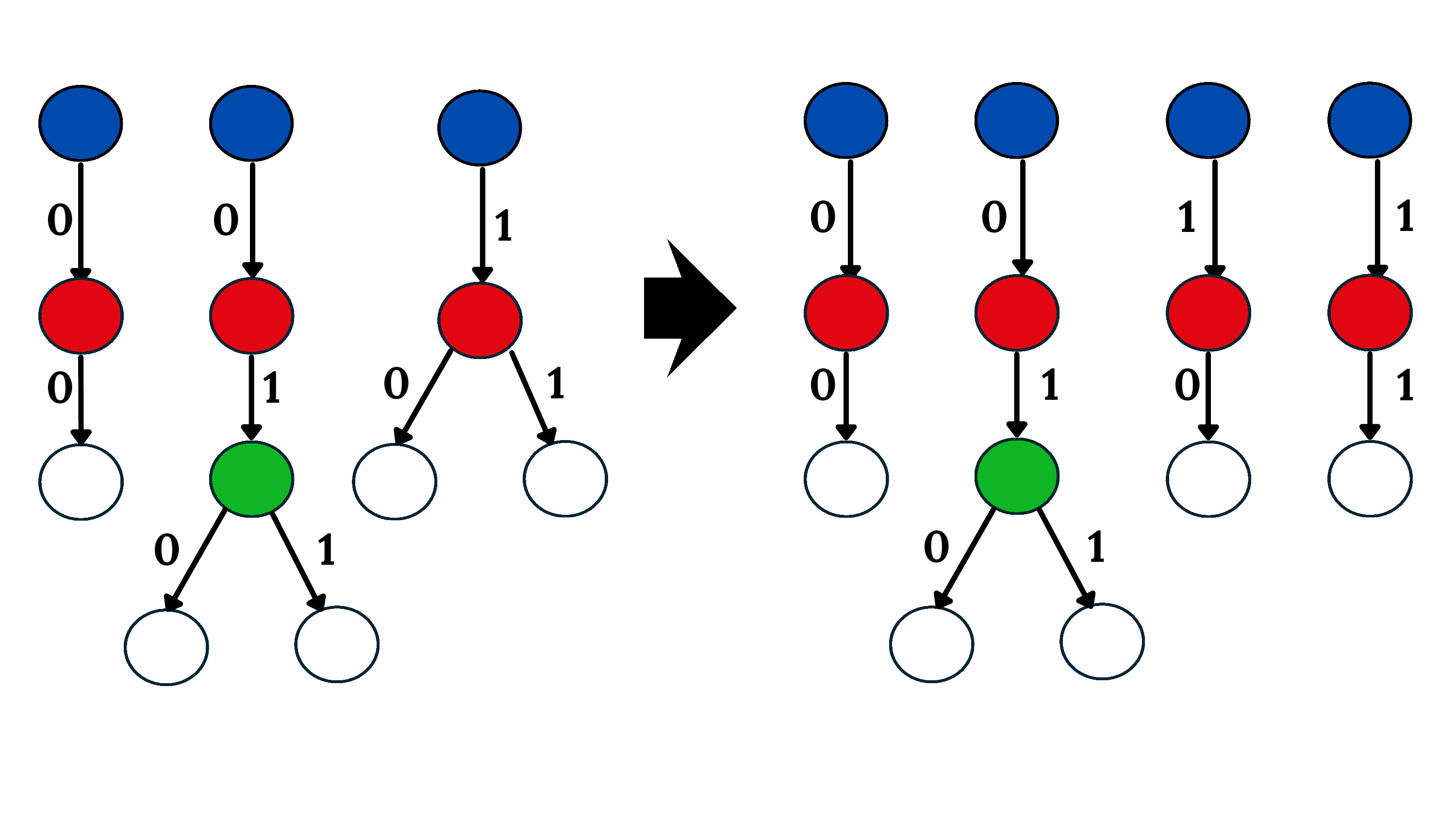}
        \caption{Desired splits at branches.}
        \label{fig:step2}
    \end{subfigure}
    \hfill
    \begin{subfigure}[t]{0.21\textwidth}
        \centering
        \includegraphics[width=\textwidth]{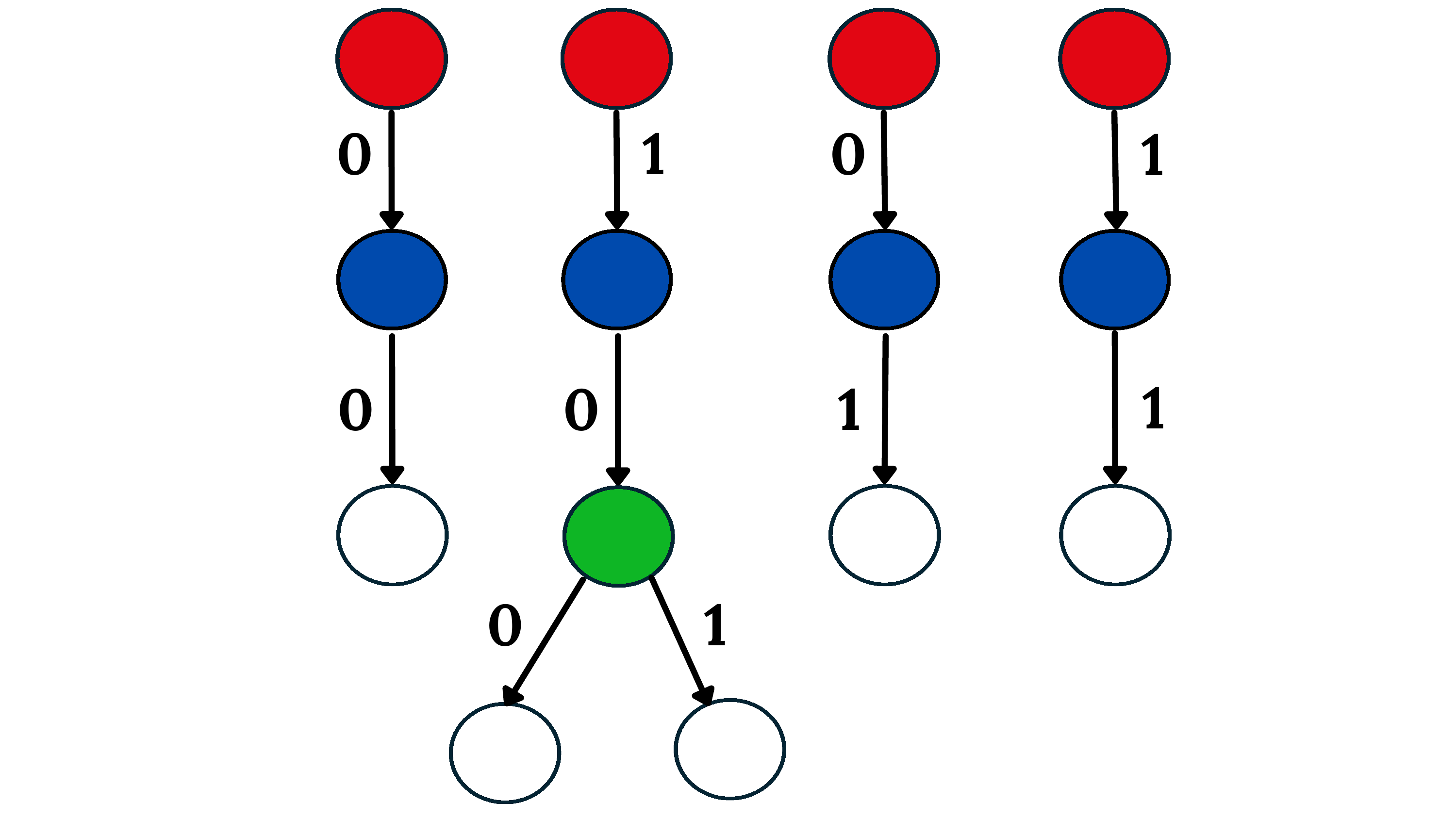}
        \caption{Move splits to the root.}
        \label{fig:step3}
    \end{subfigure}
    \hfill
    \begin{subfigure}[t]{0.24\textwidth}
        \centering
        \includegraphics[width=\textwidth]{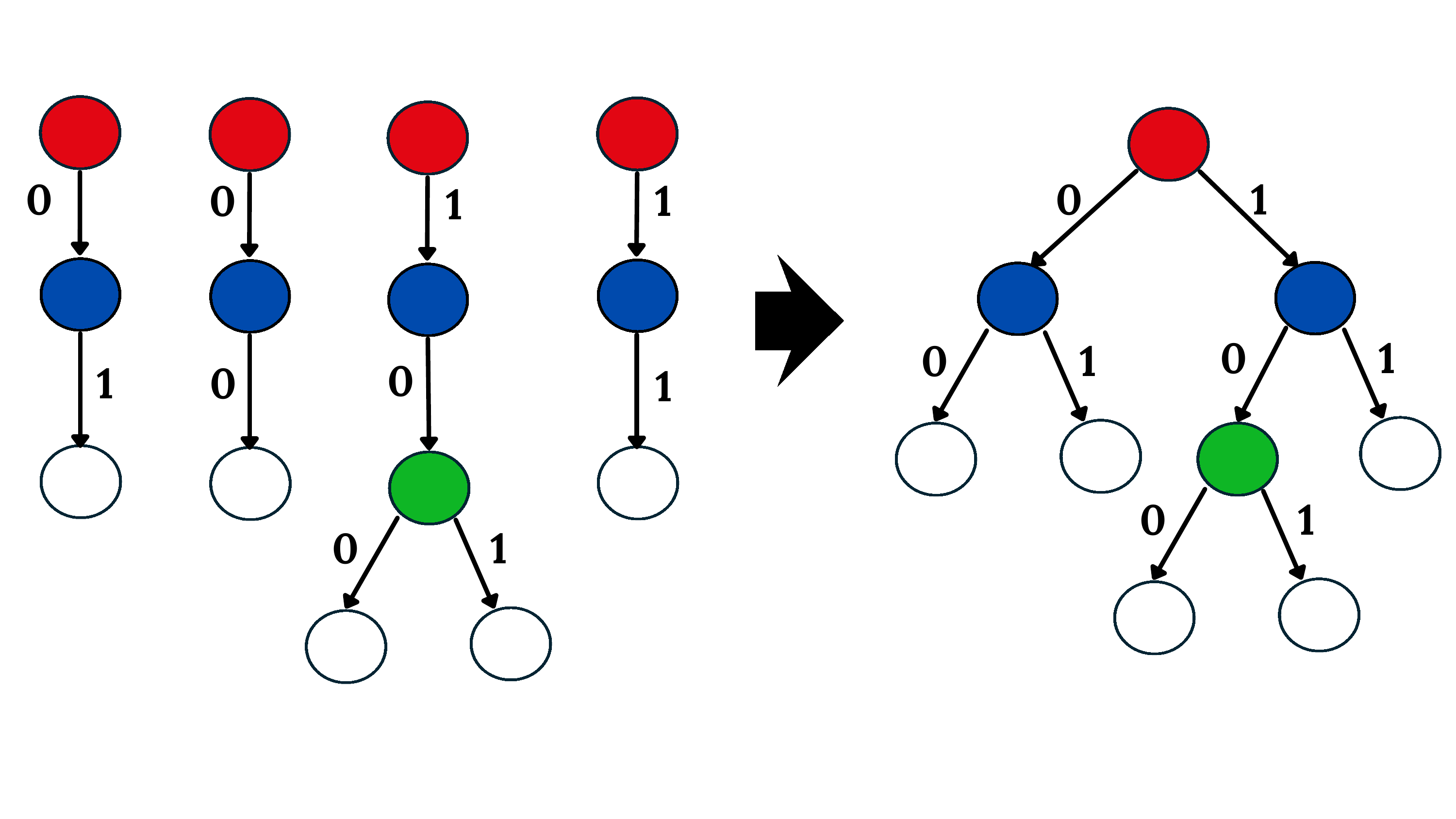}
        \caption{Iteratively rebuild the subtree.}
        \label{fig:step4}
    \end{subfigure}
    \caption{Restructuring IDTs by their intrinsic non-overlapping rule decomposition covering the full space.}
    \label{fig:restructuring}
\end{figure}

In streaming settings, models may lose not only past knowledge but also the capacity to learn new concepts. This occurs when parameters drift into regions of the loss landscape where optimization becomes slow, leading to reduced plasticity \cite{pascanu2021study}. Under this view, while plasticity is often framed as freeing capacity in IDTs, e.g., via change detectors and post-pruning \cite{nowak2025behavioral,manapragada2018extremely}, a more precise view lies in how well the current parameters serve as a starting point for further learning, independent of how much knowledge they store. In this regard, IDTs inherently exhibit low plasticity due to their conservative, history-dependent structure. Because they grow by making locally optimal splits (e.g., via thresholds, grace periods, tie-breaking rules), their ability to revise earlier decisions is limited. However, recent work has challenged this rigidity. PLASTIC \cite{heyden2024leveraging} introduces a restructuring mechanism that allows an IDT to revisit and modify pruned subtrees (Fig. \ref{fig:restructuring}), exploiting the fact that a tree’s structure can change without altering its predictive semantics. Similarly, DFDT \cite{schreckenberger2020restructuring} proposes reordering and pruning operations to promote informative attributes toward the root, enabling adaptation in trapezoidal data streams. DCFHT \cite{zhao2025online} extends this to capricious data streams.

Conversely, in neural networks, reduced plasticity often manifests as growing weight magnitudes and saturated units. Regularization-based stability methods such as EWC, MAS, and SI encourage parameters toward zero, but can unintentionally collapse weight matrix ranks and hinder adaptation. To retain plasticity, one can instead regularize toward initialization (preserving “how to learn”) \cite{kumar2023maintaining} or toward curvature-preserving parameter distributions, for example via Wasserstein-based order-statistic regularization \cite{lewandowski2023curvature}. Another direction focuses on reintroducing flexibility: S$\&$P \cite{ash2020warm} combines weight decay with stochastic perturbations to restore movement in parameter space, though at the cost of increased forgetting. ReDo \cite{sokar2023dormant} improves stability by resetting only saturated units, yet still struggles with signal propagation issues \cite{lyle2023understanding}. Continual Backprop \cite{dohare2021continual} makes this selective reset more principled by tracking utility scores over incoming and outgoing weights and protecting newly reset units until they mature. UPDG generalizes this by coupling gradient updates with adaptive perturbations, applying minimal changes to useful units and stronger rejuvenation to dormant ones \cite{elsayed2024addressing}.

\subsection{Diversification}

\begin{figure}[ht]
    \centering
    \begin{subfigure}[t]{0.21\textwidth}
        \centering
        \includegraphics[width=\textwidth]{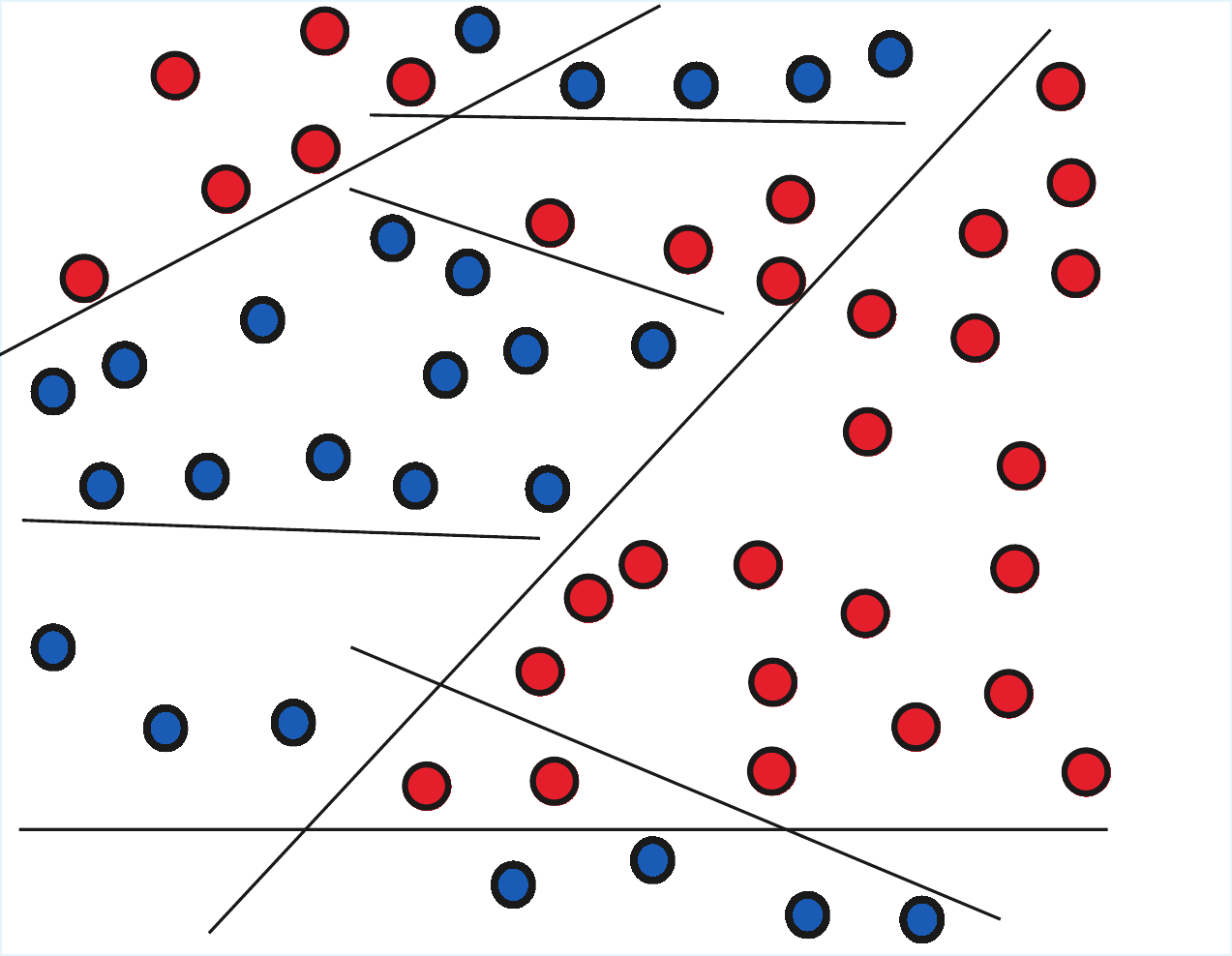}
        \caption{In a complex space.}
        \label{fig:diversification1}
    \end{subfigure}
    \hfill
    \begin{subfigure}[t]{0.25\textwidth}
        \centering
        \includegraphics[width=\textwidth]{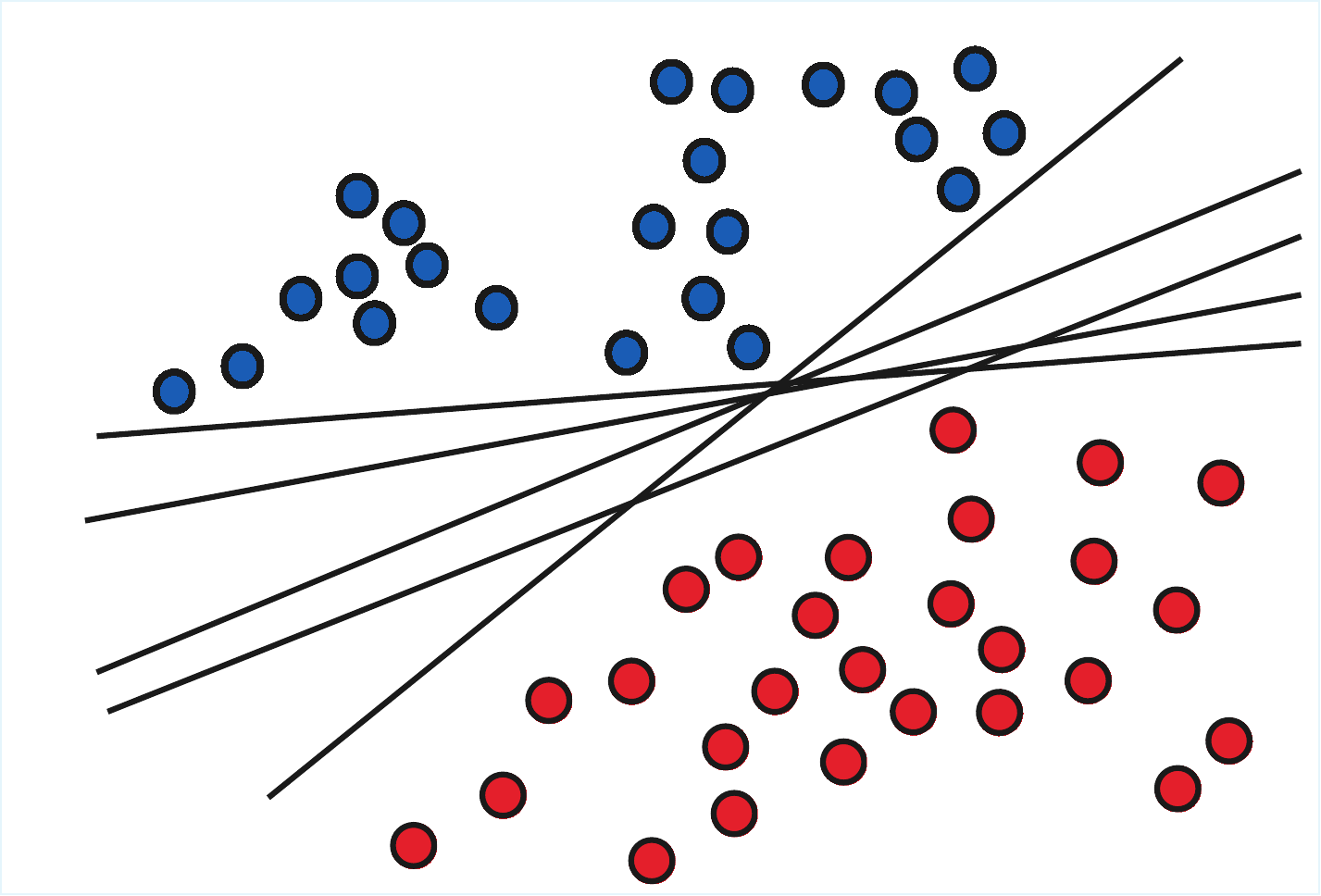}
        \caption{In a simpler space.}
        \label{fig:diversification2}
    \end{subfigure}
    \caption{Multiple non-redundant perspectives of the data.}
\end{figure}

Under tight memory constraints, a divide-and-conquer strategy naturally calls for diversification, ensuring that models store complementary rather than redundant information (Fig. \ref{fig:diversification1}). Even in cases where the data is relatively simple, one can explore different perspectives of the same patterns within a given computational budget (Fig. \ref{fig:diversification2}). This can be imposed through hard boundaries, such as explicit output specialization \cite{neves2025online}. More commonly, however, streaming ensemble methods rely on softer mechanisms that perturb the input or feature space: horizontally, e.g. through Poisson-based instance weighting \cite{oza2001experimental}, or selective instance filtering \cite{idrees2020new}; and vertically, e.g. through random subspace selection \cite{gomes2019streaming}. Beyond implicit diversification, several approaches explicitly manage the ensemble repository using diversity metrics such as double-fault \cite{abadifard2023dyned}, or the kappa statistic \cite{cano2020kappa}. At a higher level, heterogeneous ensembles can be maintained via local search heuristics \cite{veloso2018self}, evolutionary strategies \cite{moya2024improving}, and meta-learning methods \cite{rossi2014metastream}.

In contrast, neural networks typically achieve diversification within a single model by controlling how representations are shared or separated. One approach is to learn domain-invariant features by sharing parameters across layers or models \cite{wang2023model}. Another method allocates distinct feature-processing pathways while minimizing discrepancies between them, avoiding explicit mapping across domains \cite{zhang2020learning}. Distillation-based methods support this by transferring feature-level knowledge from a classifier trained on past labels to one trained on new labels, enabling the model to adapt without retraining from scratch \cite{wang2020deep}. 

\subsection{Retrieval}

\begin{figure}[ht]
    \centering
    \includegraphics[width=0.45\textwidth]{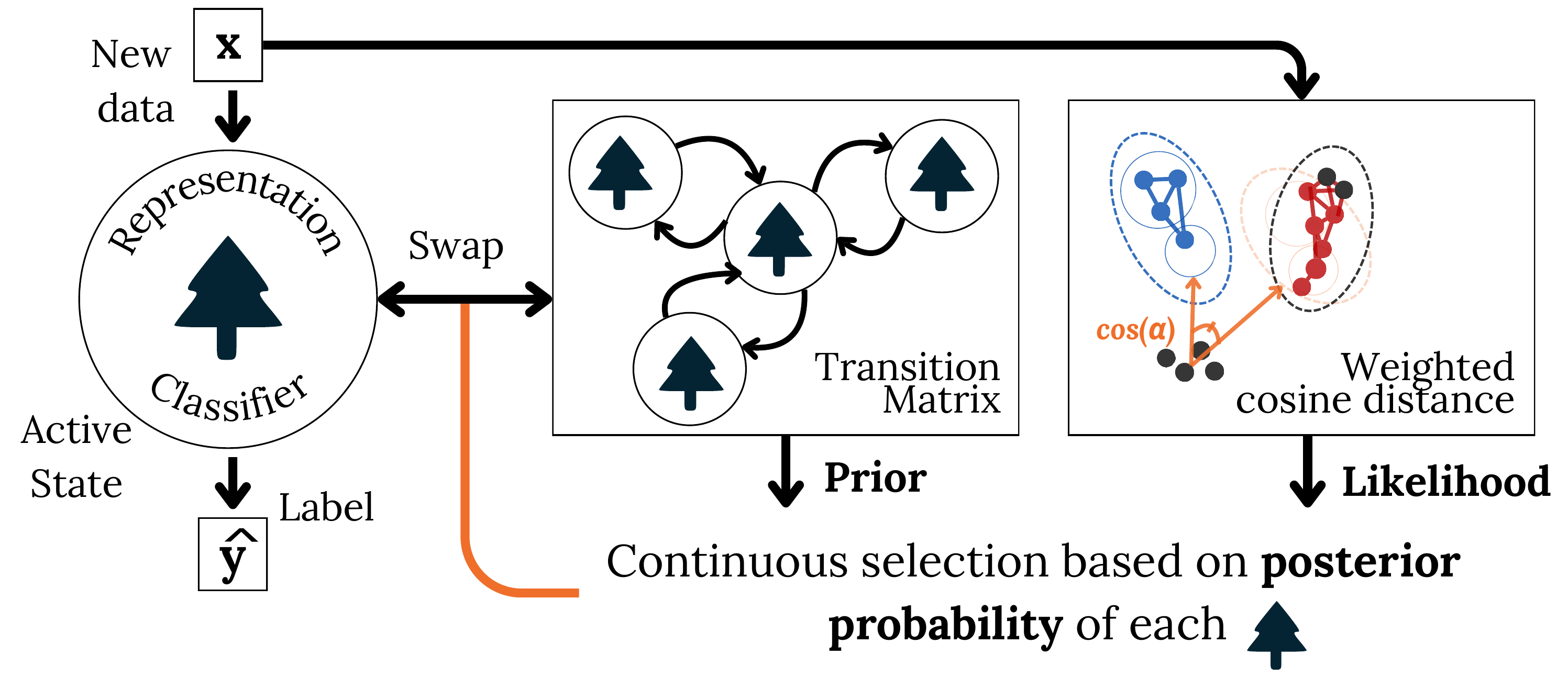}
    \caption{Posterior probabilities for all candidate models, with signals of priors (when a state is expected to reappear) and likelihoods (how much a state matches current data).}
    \label{fig:select}
\end{figure}

Balancing stability and plasticity in a shared representation is NP-hard \cite{knoblauch2020optimal}. Instead, a model must quickly recall past knowledge and decide which modules to update within a compositional framework \cite{tajwar2021no,risca2025boosting}. In SL, this is achieved in several ways. Neighborhood-based dynamic selection identifies supervised models with high competence in the local region around the query \cite{davtalab2024scalable}. Referee meta-models detect recurrence without statistical comparisons across all stored models \cite{gama2014recurrent}. Sequence mining meta-models capture patterns in stored models, relying on expectations of transitions in their competence \cite{yang2006mining, wu2022probabilistic}. Repository matching approaches reuse learned models when similar contexts reemerge, following drift detection (Fig. \ref{fig:idts2}) \cite{goncalves2013rcd}. Finally, hybrid probabilistic methods use Bayesian inference to compute posterior probabilities for all candidate states: likelihoods are obtained by comparing a meta-representation vector of each classifier to incoming data windows using weighted cosine distance, while priors are estimated via transition matrices informed by drift detectors (Fig. \ref{fig:select}) \cite{halstead2022probabilistic}. In unsupervised scenarios, novelty detection methods rely on clustering structures to find cohesive agglomeration of anomalies \cite{faria2013novelty,pauperio2025explainable}.

\begin{figure}[htbp]  
    \centering
    \begin{subfigure}[t]{0.25\textwidth}
        \centering
        \includegraphics[width=\textwidth]{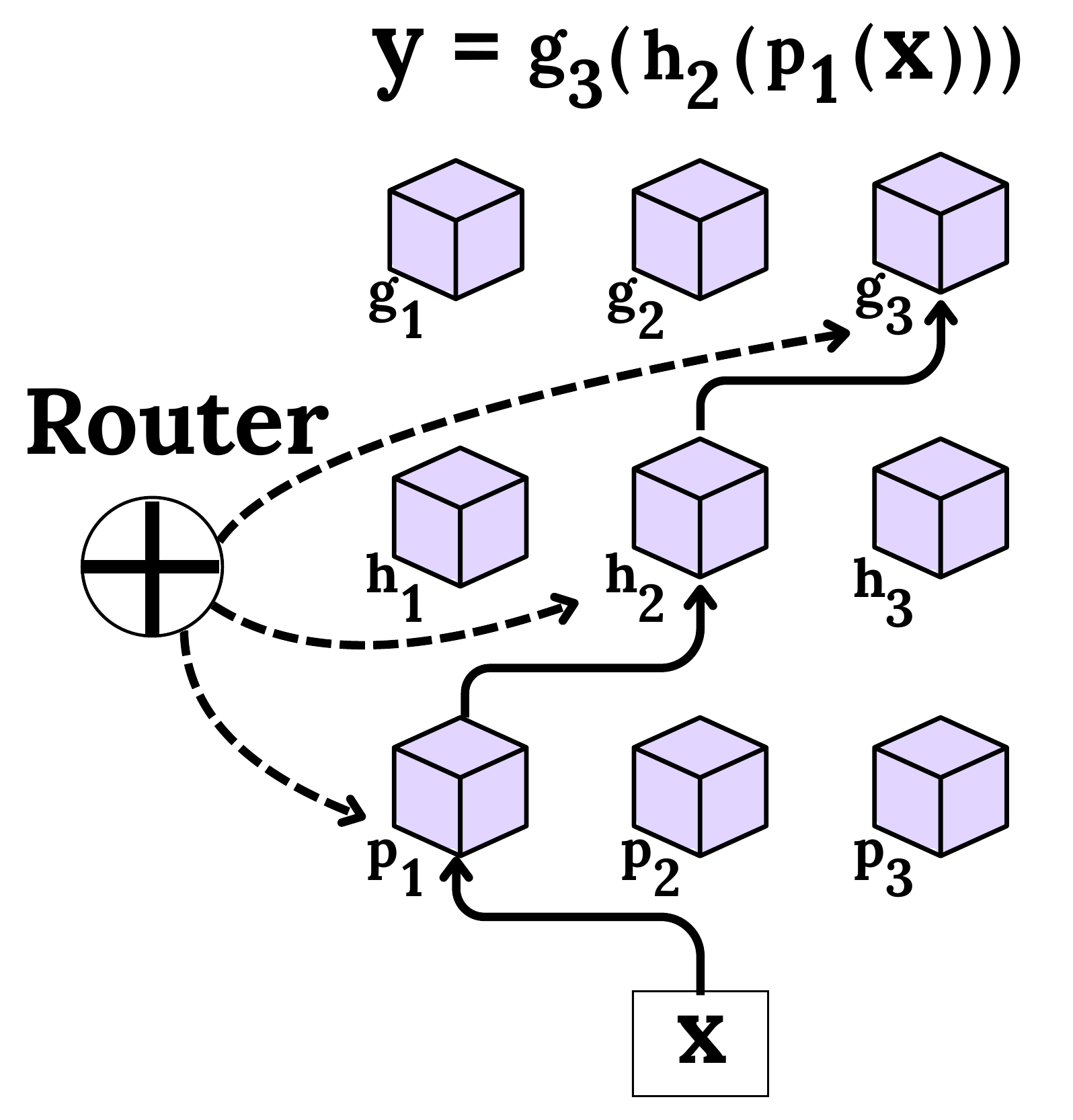}
        \caption{Modular routing.}
        \label{fig:routing}
    \end{subfigure}
    \begin{subfigure}[t]{0.21\textwidth}
        \centering
        \includegraphics[width=\textwidth]{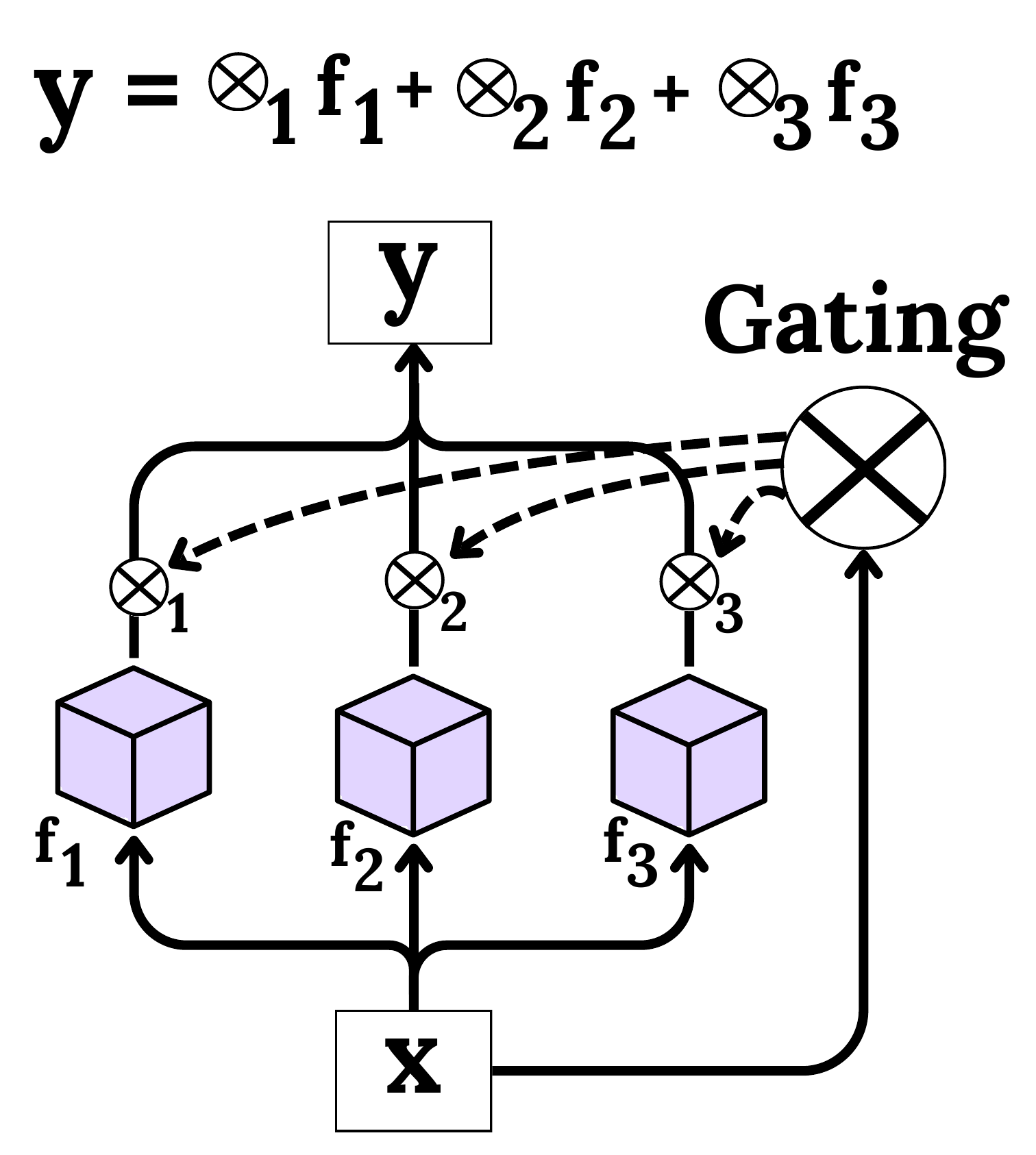}
        \caption{Mixture-of-experts.}
        \label{fig:gating}
    \end{subfigure}
    \caption{End-to-end differentiable retrieval in CL.}
\end{figure}

In neural networks, retrieval can be facilitated through end-to-end optimization. A common approach involves using multiple expert branches that are selectively activated by a gating mechanism, with their outputs integrated via a data-dependent weighting scheme (Fig. \ref{fig:gating}) \cite{shazeer2017outrageously}. Additionally, routing-based techniques allow for greater specialization by decomposing sub-concepts into sequential or parallel processing stages, enabling modules to be flexibly reused across different contexts (Fig. \ref{fig:routing}) \cite{rosenbaum2019routing, ostapenko2022attention}. In challenging unsupervised streaming scenarios, retrieval often relies on identifying latent concept boundaries. This can be achieved through methods such as cross-concept class discrimination \cite{guo2023dealing}, adversarial one-class classification \cite{sabokrou2018adversarially}, and reverse distillation from one-class embeddings \cite{deng2022anomaly}. Furthermore, cross-concept identification can be explicitly modeled using separate networks \cite{von2019continual}, learned binary masks \cite{wortsman2020supermasks}, or various out-of-distribution detection approaches \cite{lee2022theoretical}.

\section{In-context stream mining for SCL}

\begin{figure}[ht]
    \centering
    \includegraphics[width=0.48\textwidth]{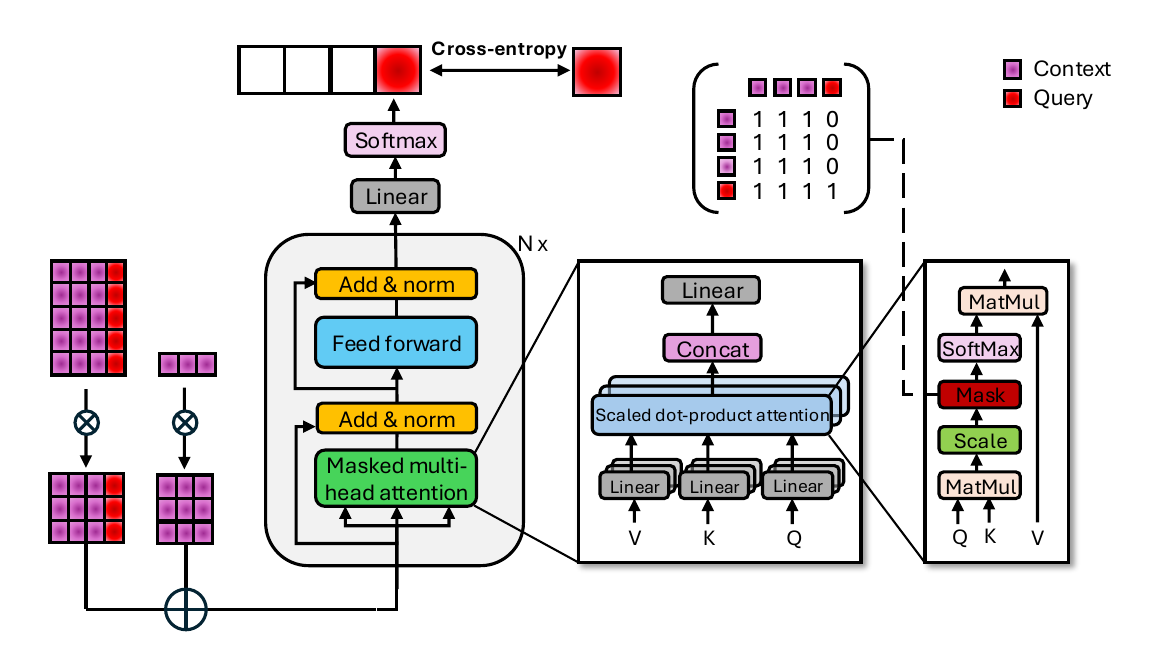}
    \caption{Exemplifying LTM's architecture and training.}
    \label{fig:architecture}
\end{figure}

Traditional SL and CL methods follow a two-stage process: optimizing models over a sliding window, followed by selecting the best configuration for the next window. Foundational models (FMs), however, enable instant model deployment, leveraging prior learning to bypass the need for extensive tuning \cite{brown2020language}. Through pretraining on vast corpora, FMs acquire soft inductive biases, drawing from a wealth of prior experiences. This results in emergent abilities, such as few-shot in-context learning (ICL), which allows models to perform new tasks during inference by conditioning on a set of input-output examples, without requiring parameter updates. Consequently, this ICL capability of FMs has spurred a new research paradigm focused on designing architectures that are pre-trained on a wide range of synthetic tabular datasets, referred to as large tabular models (LTMs) \cite{van2024tabular}. Unlike traditional models, LTMs perform instant classification without fine-tuning \cite{hollmann2025accurate, qu2025tabicl}. They adapt to unseen datasets in a single forward pass by using various training examples as context, similarly to how large language models (LLMs) use preceding tokens. Practically, an LTM is a transformer model (Fig. \ref{fig:architecture}), trained on data simulated with Bayesian neural networks or structural causal models \cite{hollmann2022tabpfn}, inductive biases from decision trees \cite{qu2025tabicl}, or DAG-based computational graphs \cite{hollmann2025accurate}, with the training set size acting as a regularizer on the network's expected complexity.

Building on these developments, a new paradigm for SL and CL centers on stream-level context construction: summarizing unbounded data streams on-the-fly before providing them to LTMs \cite{lourencco2025context}. Modern LTMs can process extremely large contexts, exceeding 500K samples \cite{qu2025tabicl} and 50K features \cite{kolberg2025tabpfn}, which provides unprecedented flexibility in how sketches of streaming data can be represented. Importantly, this data-centric perspective still allows us to leverage the aforementioned core insights from model-centric SL and CL research concerning \textbf{stability}, \textbf{plasticity}, \textbf{diversification}, and \textbf{retrieval}. In fact, these factors can now be controlled directly through context design rather than complex architectural or optimization interventions, enabling simpler and more explicit trade-offs.

Drawing from the literature on how FMs apply data selection strategies for pre-training, instruction-tuning, alignment, and in-context learning \cite{albalak2024survey}, we identify two complementary axes for organizing context design. The first concerns \textbf{distribution matching}: selecting data similar to the target, yielding \textbf{plasticity} when emphasizing the current distribution, and \textbf{stability} when maintaining support across prior distributions. The second concerns \textbf{distribution compression}: reducing redundancy while maintaining representational power, which supports \textbf{diversification} when filling memory with non-overlapping representative samples, and \textbf{retrieval} when dynamically constructing a task-specific context from a larger pool.

\subsection{Distribution matching}

\begin{figure}[ht]
    \centering
    \includegraphics[width=0.47\textwidth]{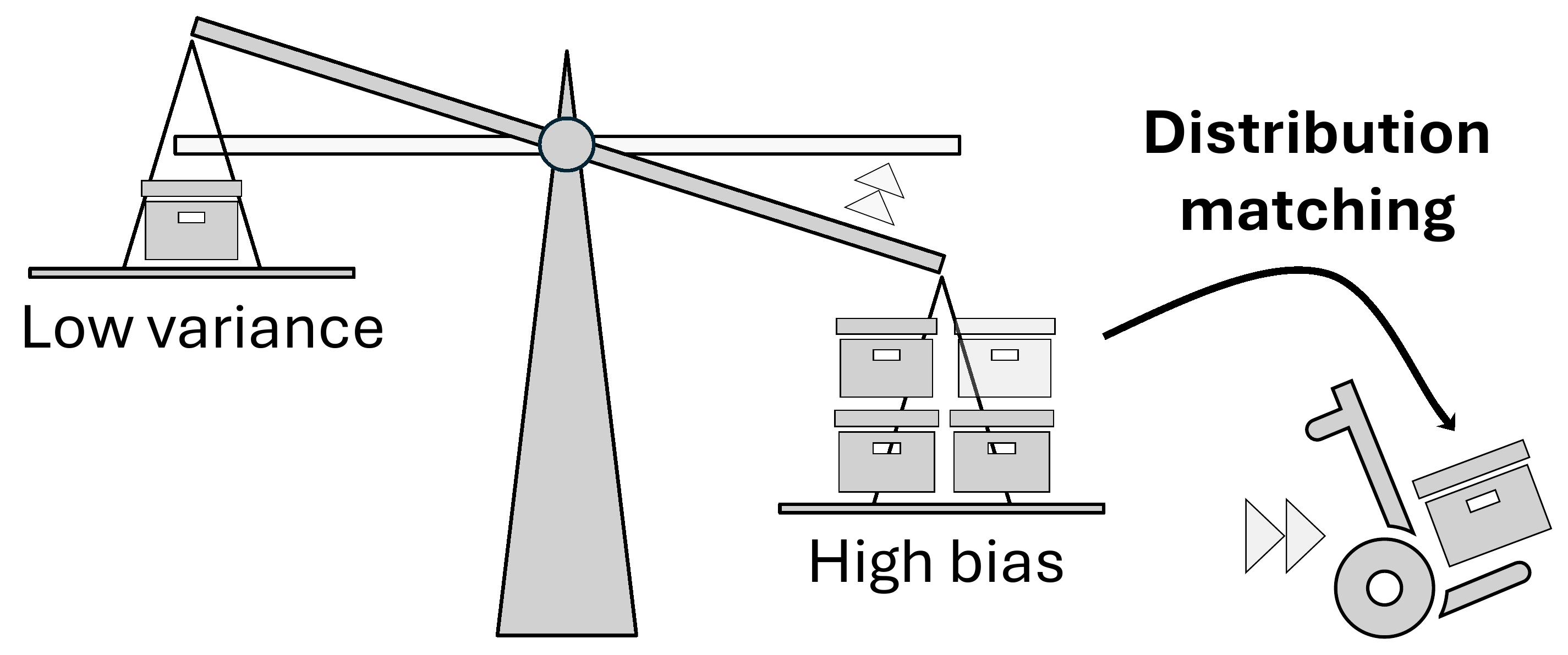}
    \caption{In-context learning from frequentist viewpoint.}
    \label{fig:frequentist}
\end{figure}

To better understand the goal of distribution matching, one can adopt a frequentist perspective (Fig. \ref{fig:frequentist}) \cite{nagler2023statistical}. From a variance standpoint, an LTM, pre-tuned but untrained, with many hyperparameters and multi-head attention, is highly sensitive to individual context samples, which increases its ability to select effective submodels and reduces predictor variance. From a bias standpoint, hyperparameters are optimized for the prior task distribution. If the prior is broad and not overly concentrated away from the true hypothesis, the posterior predictive distribution closely approximates the true predictive distribution. Consequently, the LTM’s ability to learn at inference depends on its structural properties, with the optimal approximation characterized by a Kullback-Leibler criterion \cite{nagler2023statistical}. Intuitively, adding more context samples reduces sensitivity to minor input perturbations, lowering variance, while bias persists unless the context is concentrated near the target distribution.

With this perspective, when the data distribution shifts over time, the context may no longer reflect the current environment, leading the model to produce biased predictions based on outdated patterns. This requires a design trade-off:

\begin{itemize}
\item \textbf{Plasticity:} prioritizing recent examples to adapt quickly to new local patterns, at the risk of losing information about past classes and concepts.
\item \textbf{Stability:} retaining examples from earlier concepts to maintain a global understanding of all classes, including those observed in the distant past.
\end{itemize}

\begin{figure}[ht]
    \centering
    \includegraphics[width=0.37\textwidth]{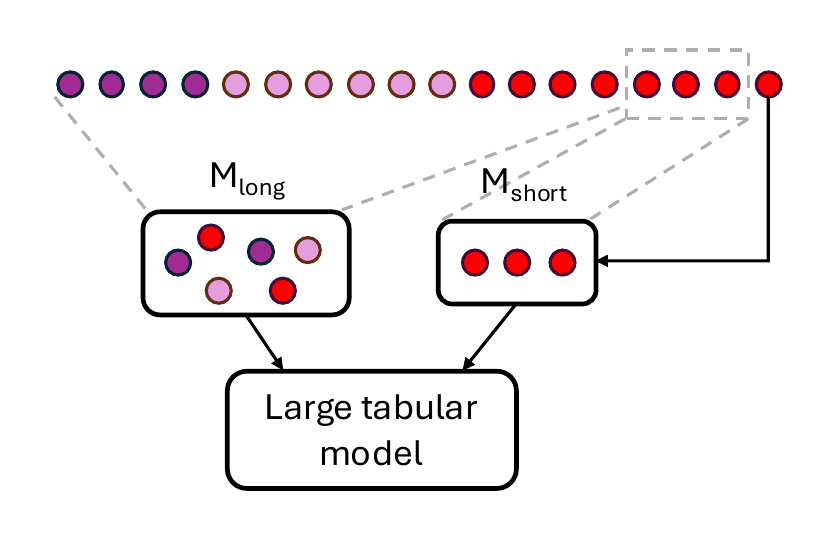}
    \caption{Short-term memory maintains recent instances. Long-term memory preserves infrequently seen classes.}
    \label{fig:naive}
\end{figure}

A simple yet effective solution is a dual-memory FIFO system  (Fig. \ref{fig:naive}) \cite{lourencco2025context}. The long-term memory stores a fixed set of older samples across all known classes, preserving rare or infrequently seen categories. In contrast, the short-term memory maintains the most recent portion of the stream, capturing local variations, transient sub-concepts, and evolving intra-class dynamics. By combining these two memories, the model achieves a balance between long-term stability and rapid adaptability to short-term fluctuations in the data distribution.

However, this approach is naive. Plasticity is largely reactive, implemented via fading strategies rather than proactive adaptation, while stability only addresses catastrophic forgetting of classes, without ensuring invariant representations across all concepts. Addressing this requires diversification and retrieval principles, adopting inductive biases to selectively match data distributions, such as: smoothness, where nearby points in high-density regions are assumed to produce similar outputs; clustering, where points in the same vicinity likely share a concept; and manifold, where high-dimensional data lies on shared low-dimensional latent representations. Notably, because in-context stream mining is formulated as a data selection problem rather than explicit model design, these inductive biases naturally align with the goal of distribution compression, i.e., promoting efficient and representative context construction.

\subsection{Distribution compression}

Distribution compression aims to prioritize heterogeneity and remove redundancies to reduce dataset size. Different communities approach this goal differently: SL uses synopses and sketches to summarize unbounded streams \cite{silva2013data}; CL relies on experience replay to retain past concepts \cite{chaudhry2019tiny}; and FMs apply data selection strategies across pre-training, instruction-tuning, alignment, in-context learning, and fine-tuning \cite{albalak2024survey}. Despite these differences, compression can generally be framed as a two-stage process:

\begin{itemize}
\item \textbf{Diversification:} populating and updating memory with representative, meaningful samples, through informed addition and deletion.
\item \textbf{(Optional) Retrieval:} distinguishing between memory population (which samples to store) and sampling (which points to use for in-context learning).
\end{itemize}

\begin{figure}[ht]
    \centering
    \includegraphics[width=0.45\textwidth]{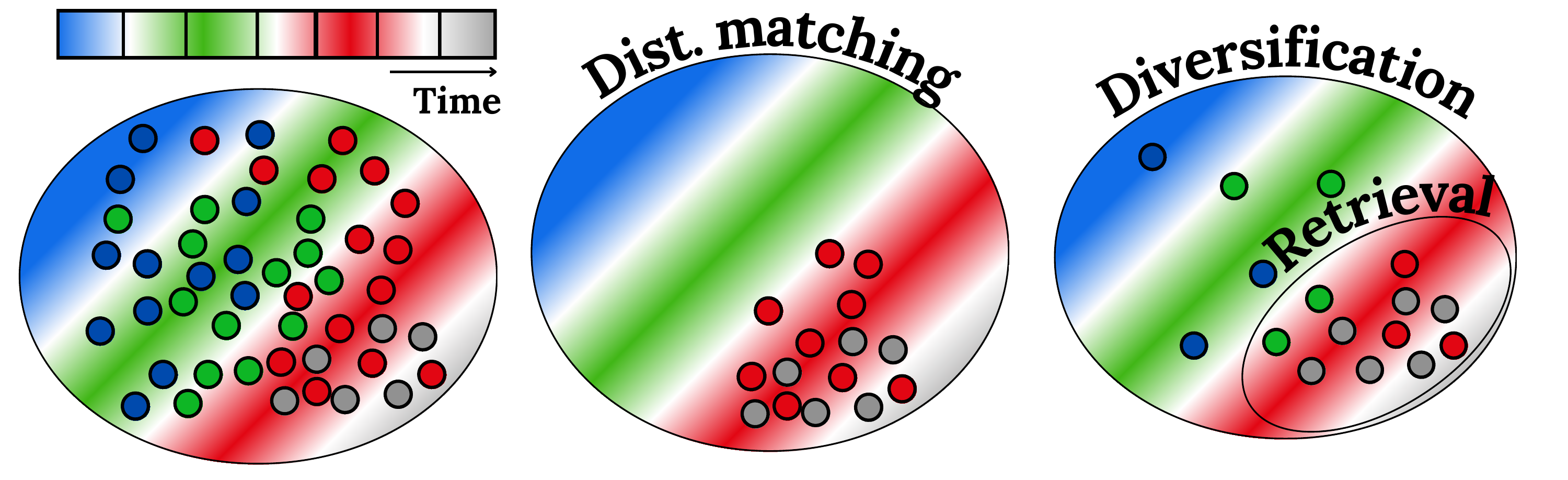}
    \caption{Distribution matching selects data similar to the target. Distribution compression maintains a diversified memory, and (optionally) retrieves data similar to the target.}
    \label{fig:diversification}
\end{figure}

\begin{figure}[ht]
    \centering
    \includegraphics[width=0.45\textwidth]{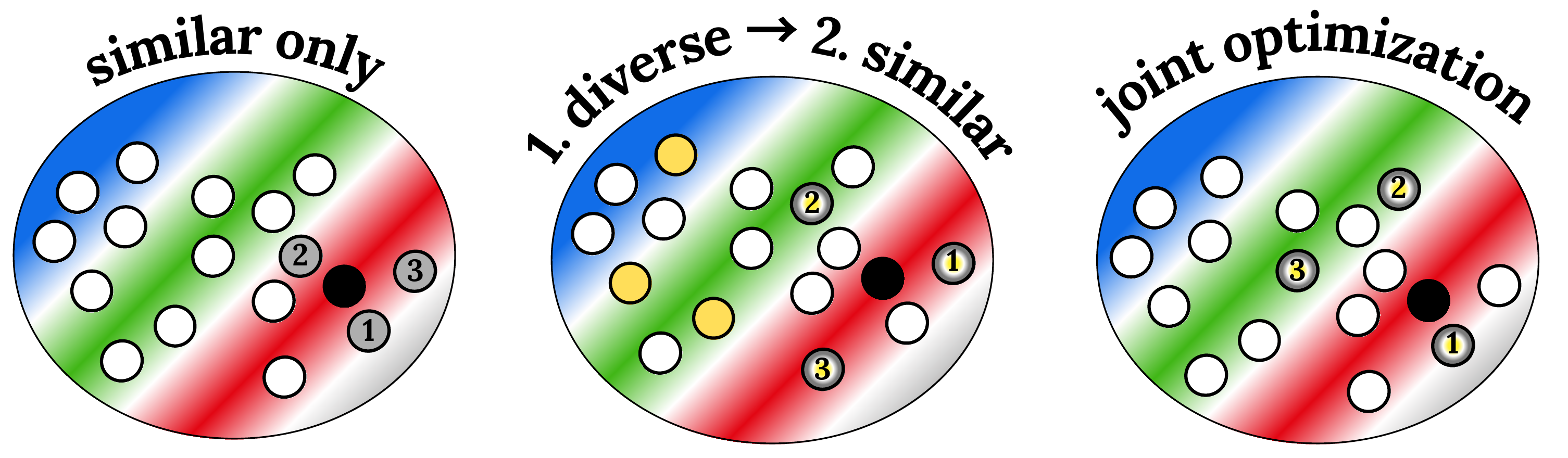}
    \caption{Selection criteria: (left) top 3 most similar from the full dataset; (middle) top 3 from a compressed set of 6 diversified instances; (right) top 3 from the full dataset using an unified metric that balances similarity and diversity.}
    \label{fig:retrieval}
\end{figure}

Distribution compression is inherently linked to distribution matching, shaping how the model balances stability and plasticity when selecting context examples. However, the relationship between diversification and these objectives is subtler than it appears: although diversification is often associated with enhanced stability, it does not necessarily compromise plasticity. For example, selecting examples solely based on similarity to the query in embedding space \cite{wu2022self} can lead to redundancy and omit less similar yet informative concepts that support contrastive learning \cite{xiao2025role}. In contrast, true distribution matching aims to maximize feature coverage \cite{levy2022diverse}, select examples according to difficulty \cite{cook2025no}, and consider sample sensitivity \cite{chen2022relation}. Empirical evidence supports this (Fig. \ref{fig:retrieval}): sequential methods that explicitly balance similarity to the query with diversity among selected examples consistently outperform naive strategies, such as choosing the $K$ most similar examples, or selecting similar examples from a diversity-reduced subset \cite{xiao2025role}.

Viewed from this perspective, diversification emerges as a mechanism that can simultaneously enhance both plasticity and stability. The benefits of diversification, however, are limited. Its effectiveness is constrained by the inherent difficulty of learning incremental concepts within a finite parameter space. In this context, retrieval serves as a complementary strategy: by separating points for new knowledge from those revisiting prior knowledge, retrieval allows for a divide-and-conquer approach \cite{wang2022dualprompt}. Importantly, retrieval itself can be understood as a higher-level application of diversification. While diversification maintains representative and informative examples in memory to support both plasticity and stability, retrieval applies the same principle to select query-specific context subsets, reducing interference between old and new concepts.

\begin{figure}[ht]
    \centering
    \includegraphics[width=0.43\textwidth]{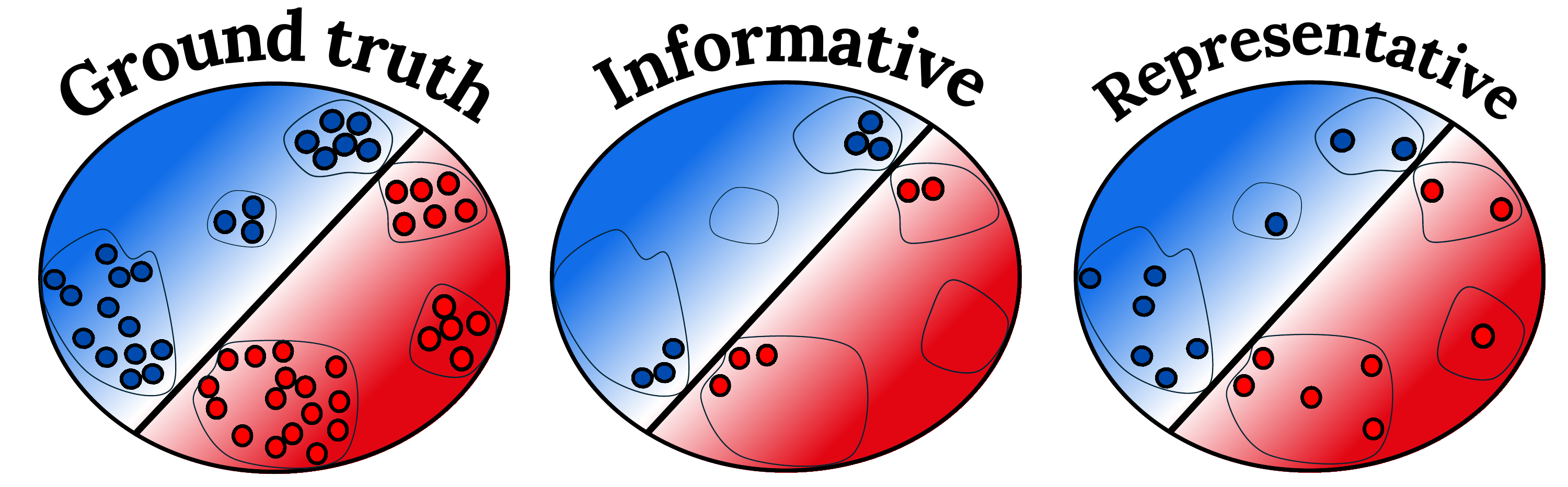}
    \caption{Selection criteria: informativeness prioritizes uncertain or difficult instances, while representativeness aims to capture the overall data distribution.}
    \label{fig:alstrategies}
\end{figure}

With this unified perspective, it becomes clear that heuristic or score-based methods benefiting diversification also enhance retrieval, they simply operate at different stages of the learning process. Critically, any such process must account for the fact that data points vary in potential: some are more representative or informative than others. Classical streaming data selection methods address this by framing the problem as online clustering, leveraging synopsis techniques such as histograms, wavelets, or sketches to construct geometric and statistical descriptors of the data \cite{silva2013data}. Concept drift is typically detected by tracking assignment errors \cite{sakamoto2015concept}, or comparing recent and reference data within clusters using tests like the univariate k-sample Anderson-Darling for each principal component of each centroid \cite{wan2021concept}. However, these approaches are largely unsupervised, whereas our focus is on predictive performance. Predictive strategies may instead maintain a dynamic set of short-term and long-term prototypes based on error-driven representativeness learning and constrained clustering inspired by synchronization \cite{shao2014prototype}. In this vein, numerous data selection methods have emerged across the fields of active learning \cite{sener2017active}, continual learning \cite{toneva2018empirical, castro2018end}, and training dynamics \cite{koh2017understanding}. These methods generally assign each instance a scalar score, either reflecting informativeness or representativeness (Fig. \ref{fig:alstrategies}) \cite{pecher2024automatic}.

On one hand, \textbf{informativeness} measures how critical a sample is for learning, often favoring hard or uncertain instances. In active learning, this includes points where the model exhibits low confidence \cite{krawczyk2017active}, high reconstruction errors \cite{yu2025ensemble} or high disagreement \cite{cavalcanti2025arm}. In continual learning, analogous metrics are employed, but with the implicit goal of mitigating catastrophic forgetting. Methods include herding selection \cite{welling2009herding,castro2018end}, which samples based on distance to the class mean; discriminative sampling \cite{liu2020mnemonics}, which targets decision boundary points; and entropy-based sampling \cite{chaudhry2018riemannian}, which selects high-uncertainty predictions. In training dynamics, sample quality is assessed via learnability heuristics such as forgetting frequency \cite{toneva2018empirical} or ease of learning \cite{swayamdipta2020dataset}. Alternatively, separate scoring models \cite{rubin2021learning, li2023finding} or reinforcement learning strategies \cite{zhang2022active, lu2022dynamic} can optimize sample selection.

On the other hand, \textbf{representativeness} captures how well a subset reflects the overall data distribution, complementing informativeness. In active learning, these help balance exploration-exploitation trade-offs, capturing the structure of the raw data \cite{himaja2024cluster, yin2023clustering} or the embedding space \cite{zhang2021nonstationary}. In contrast, continual learning faces the challenge of maintaining memory samples, where similarity is often be balanced with diversity \cite{sun2022information, wiewel2021entropy, qin2023context}, or informativeness with representativeness \cite{bang2021rainbow, levy2022diverse}. Examples include CoPE \cite{de2021continual}, which maintains class prototypes in a shared latent space while minimizing intra-class variance and maximizing inter-class separation, and core-set methods such as cardinality-constrained bilevel optimization \cite{borsos2020coresets, killamsetty2021glister, mirzasoleiman2020coresets}. However, unlike standard core-set selection, in-context stream mining does not involve parameter updates.

All these principles naturally extend to in-context stream mining, with one key distinction: in LTMs, the memory population effectively is the model, since it directly defines the implicit decision boundaries that the LTM can express \cite{lourencco2025context}. In contrast, in experience replay-based CL, the memory serves primarily as a support mechanism for a separate parametric model, where samples are selectively replayed to prevent forgetting, typically chosen using criteria such as max loss, min margin, min logit-distance, or min confidence \cite{prabhu2020gdumb}.

From this perspective, the statistical uncertainty of in-context stream mining is primarily determined by which observations are missing or underrepresented in the memory. While conventional stream mining emphasizes accumulated prequential model uncertainty, in-context stream mining focuses on how the current query would shift in the hypothesis space in response to a particular context-query pairing. This is analogous to transductive reasoning, where predictions are made on a closed set of instances without constructing a general model.

Thus, while heuristic or score-based methods remain useful for evaluating individual samples, in-context stream learning benefits from assessing the joint influence of sample subsets. Inclusion or removal of combinations should be evaluated by their collective impact on learning \cite{yang2022dataset}, which can be operationalized by prompting the LTM to rate samples and observing performance changes, analogous to a leave-one-out procedure \cite{tan2023data}. However, directly computing the performance drop for every possible instance not only may provide insufficient signal, but also is computationally infeasible, requiring $2^n$ inferences for a dataset of size $n$. Nonetheless, the framework is valuable for studying how combinations of training examples, rather than individual ones, affect generalization, for example through inclusion or exclusion of entire prototypical classes \cite{jain2023data}.

To make this approach tractable in practice, data selection can be framed as an online learning-to-rank (LTR) problem, which leverages the counterfactual effects of sample inclusion while minimizing LTM calls \cite{grotov2016online}. This naturally aligns with reinforcement learning or contextual bandit frameworks, where the system selects ranked lists of exemplars and receives rewards based on observed performance \cite{purohit2025sample}. The core challenge is balancing exploration (testing new rankings) with exploitation (using the best-known ranking).

\section{Conclusions}

LTMs emerge as a promising bridge between the CL and SL communities, warranting further investigation. While extreme real-time and edge constraints still requires smaller and faster LTMs, we set aside such engineering constraints, which are likely to be addressed by advances in the TinyML community \cite{somvanshi2025tiny}. Our focus is instead on the SCL context, where the central challenge is not model size or speed, but orchestrating the dynamic interplay of data arrival, training, recovery, and inference. As LTMs naturally become more efficient, research should prioritize algorithms that embody the core principles of streaming continual learning: stability (preserving past knowledge), plasticity (adapting to new data), diversification (reducing redundancy), and retrieval (enabling faster remembering). To illustrate the practical realization of these principles, we draw on insights from multiple communities: from SL, synopsis and sketches provide effective stream summarization \cite{silva2013data}; from CL, experience replay preserves past concepts \cite{chaudhry2019tiny}; and from foundational models, data selection strategies enhance pre-training, fine-tuning, and in-context learning \cite{albalak2024survey}.

Looking further ahead, end-to-end solutions could selectively activate LTM components per instance, guided by specialized prompter contexts fine-tuned on the most relevant data \cite{xu2024mixture} This strategy is already supported by evidence that fine-tuning improves retrieval-based performance \cite{breejen2023fine, thomas2024retrieval}. However, applying such strategies in continual learning remains a significant challenge across all modalities of foundational models \cite{bell2025future,coleman2025parameter}. We therefore encourage the community to begin exploring data selection strategies as lightweight inference-time wrappers, which can guide future developments in a manner similar to the evolution of prompt engineering in NLP and vision. The authors also plan to pursue some of these directions and welcome feedback and collaboration from those interested in contributing to this line of research.

\section*{Acknowledgments}

Funded under Ph.D. scholarship PRT/BD/154713/2023 and project doi.org/10.54499/UIDP/00760/2020 by FCT.

\bibliography{aaai2026}

\end{document}